%% file: main.tex
\documentclass[conference]{IEEEtran}
\usepackage[numbers]{natbib}
\usepackage{graphicx}
\usepackage{dirtytalk}
\usepackage{etoolbox}
\usepackage{booktabs}
\usepackage{multirow}

\usepackage[table,xcdraw]{xcolor}
\makeatletter

\patchcmd{\@makecaption}
 {\\}
 {.\ }
 {}
 {}

\ifCLASSINFOpdf
\else
\fi
\usepackage{amsmath}
\usepackage{amssymb}
\usepackage{threeparttable}

\newcommand{\figref}[1]{Fig.\hspace{1mm}\ref{#1}}
\newcommand{\tabref}[1]{Table\hspace{1mm}\ref{#1}}

\ifCLASSOPTIONcompsoc
\usepackage[caption=false,font=normalsize,labelfont=sf,textfont=sf]{subfig}
\else
\usepackage[caption=false,font=footnotesize]{subfig}
\fi
\begin{document}

\title{A comparable study: Intrinsic difficulties of practical plant diagnosis from wide-angle images}
\author{\IEEEauthorblockN{Katsumasa Suwa\IEEEauthorrefmark{1}, Quan Huu Cap\IEEEauthorrefmark{1}, Ryunosuke Kotani\IEEEauthorrefmark{1}, Hiroyuki Uga\IEEEauthorrefmark{2}, Satoshi Kagiwada\IEEEauthorrefmark{3}, Hitoshi Iyatomi\IEEEauthorrefmark{1}}
\IEEEauthorblockA{Email: kt.suwa1023@gmail.com\quad huu.quan.cap.78@stu.hosei.ac.jp\quad ryunosuke.kotani.58@hosei.ac.jp \\uga.hiroyuki@pref.saitama.lg.jp\quad kagiwada@hosei.ac.jp\quad iyatomi@hosei.ac.jp}
\IEEEauthorblockA{\IEEEauthorrefmark{1}Applied Informatics, Graduate School of Science and Engineering, Hosei University, Tokyo, Japan}
\IEEEauthorblockA{\IEEEauthorrefmark{2}Saitama Agricultural Technology Research Center, Saitama, Japan}
\IEEEauthorblockA{\IEEEauthorrefmark{3}Clinical Plant Science, Faculty of Bioscience and Applied Chemistry, Hosei University, Tokyo, Japan}}

%

\maketitle
\begin{abstract}
    \input{00_abstract.tex}
\end{abstract}


\section{Introduction}
    \input{01_introduction.tex}

\section{Materials and Methods}
    \input{02_method.tex}

\section{Experimental results}
    \input{03_experiment.tex}

\section{Discussion}
    \input{04_discussion.tex}

\section{Conclusion}
    \input{05_conclusion.tex}

\section*{Acknowledgment}
This research was partially supported by the Ministry of Education, Culture, Science and Technology of Japan (Grant in Aid for Fundamental research program (C), 17K8033, 2017-2020).



%
\nocite{*}
\footnotesize{
\bibliographystyle{IEEEtran}
\bibliography{reference}
}
\end{document}

%% file: 00_abstract.tex
Practical automated detection and diagnosis of plant disease from wide-angle images (i.e. in-field images containing multiple leaves using a fixed-position camera) is a very important application for large-scale farm management, in view of the need to ensure global food security. 
However, developing automated systems for disease diagnosis is often difficult, because labeling a reliable wide-angle disease dataset from actual field images is very laborious. 
In addition, the potential similarities between the training and test data lead to a serious problem of model overfitting. 
In this paper, we investigate changes in performance when applying disease diagnosis systems to different scenarios involving wide-angle cucumber test data captured on real farms, and propose an effective diagnostic strategy. 
We show that leading object recognition techniques such as SSD and Faster R-CNN achieve excellent end-to-end disease diagnostic performance only for a test dataset that is collected from the same population as the training dataset (with F1-score of 81.5\% -- 84.1\% for diagnosed cases of disease), but their performance markedly deteriorates for a completely different test dataset (with F1-score of 4.4 -- 6.2\%). 
In contrast, our proposed two-stage systems using independent leaf detection and leaf diagnosis stages attain a promising disease diagnostic performance that is more than six times higher than end-to-end systems (with F1-score of 33.4 -- 38.9\%) on an unseen target dataset. 
We also confirm the efficiency of our proposal based on visual assessment, concluding that a two-stage model is a suitable and reasonable choice for practical applications.

%% file: 01_introduction.tex
\begin{figure*}[hbt!]
\centering
\includegraphics[width=0.98\linewidth]{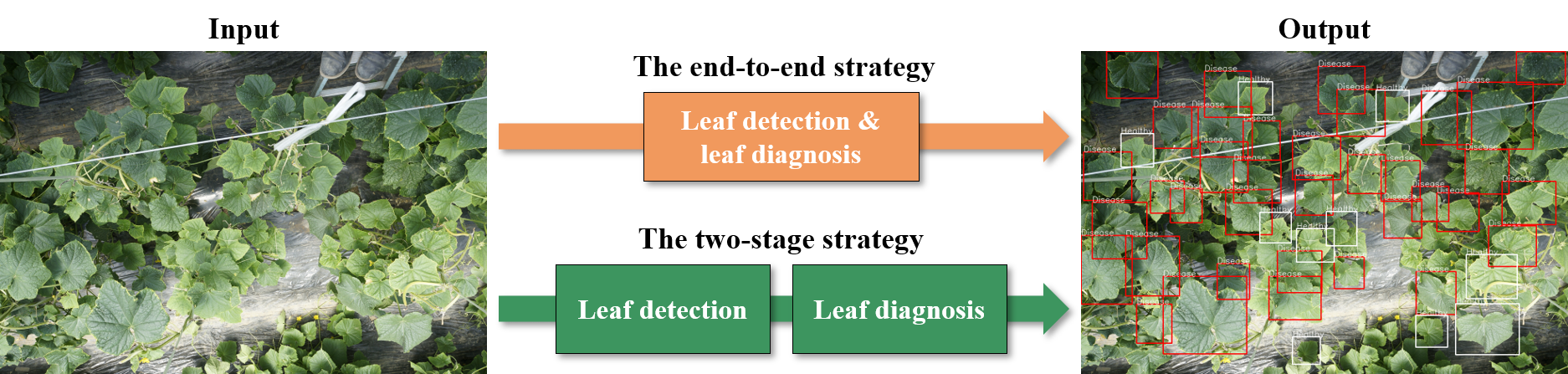}
\caption{
    Overview of the end-to-end and two-stage strategies (red and white boxes indicate disease and healthy leaves, respectively).
}
\label{fig:the_overview_system}
\end{figure*}

Plant diseases represent an enormous challenge to worldwide agricultural production, as they account for huge crop losses and pose a threat to global food security. 
It is known that losses caused by plant diseases account for at least 10\% of global food production~\cite{strange2005plant}. 
In addition, the United Nations Food and Agriculture Organization (FAO) has reported that the world’s population is expected to grow to almost 10 billion by 2050~\cite{fao2017}, which will boost demand for agricultural productivity. 
Early detection and appropriate treatment are therefore essential in order to minimize the damage caused by plant diseases and to ensure the yield and quality of global food production.

In an agricultural context, the traditional method of plant disease detection is visual observation by experts, and genetic testing is applied if necessary, requiring a great deal of specialist experience and knowledge. 
Thus, diagnosing plant diseases on real farms is usually time-consuming and expensive. 
A timely and accurate method of plant diseases recognition is in great demand.

In recent years have seen the successful application of a deep learning method called convolutional neural network (CNN) in many areas of industry. 
Due to the outstanding performance of CNNs, the use of computer vision in the field of precision agriculture has also been expanded, as it has proven to be a good option for plant disease recognition. 

Some of the early studies performed disease diagnosis based on single-image inputs (i.e. images that typically contain one single object). 
Mohanty et al.~\cite{mohanty2016} performed disease recognition for 14 crop species and 26 diseases on the PlantVillage~\cite{hughes2015open} open dataset, using pre-trained deep learning models. 
The highest classification accuracy was 99.4\% on a held-out test set. 
Liu et al.~\cite{liu2018identification} collected their own apple leaf dataset that included four common diseases, and their CNN model successfully classified apple diseases with an overall accuracy of 97.6\%. 
Despite showing excellent performances, the main drawback of these studies was that those datasets were taken from an experimental (laboratory) setup, under relatively ideal conditions for machine perception. 
That is, all the leaves were cropped beforehand, and each leaf was photographed against a uniform background, rather than under the real conditions in the cultivation field. 
Thus, the diagnostic performance was biased. 
The authors themselves stated that the accuracy of the system greatly decreased when applied to real images from the cultivation field.

In order to develop diagnostic systems for practical environments, several deep learning-based techniques have been also investigated. 
Kawasaki et al.~\cite{kawasaki2015} trained a simple but effective CNN model for the in-field classification of two cucumber diseases and healthy leaves. 
In another study of cucumber disease, Fujita et al.~\cite{fujita2016} customized a pre-trained VGG-net~\cite{vgg2014} to classify seven types of cucumber disease and one healthy from on-site cucumber leaf images under various photographic conditions. 
They reported an average accuracy of 82.3\% using a 4-fold cross-validation strategy. 
Tani et al.~\cite{hiroki2018diagnosis} designed a smaller version of the VGG-net that could classify multiple-disease infections. 
Their system attained an average accuracy of 95.0\% in the classification of 13 combinations of infections with multiple diseases on a practical cucumber dataset. 
Ramcharan et al.~\cite{ramcharan2019mobile} developed a cassava leaf diagnosis system that was available for ordinary mobile phones, and achieved an average accuracy of 80.6\% for different stages of three diseases. 
Even though the proposals mentioned above were practical, one major issues with these studies is their inconvenience for application to large-scale field images, since these systems accept only single-leaf input images.

Several sophisticated end-to-end systems from wider-area agricultural images have been proposed to alleviate this practical problem, and most of these techniques are based on recent, powerful, CNN-based real-time object detection methods such as Faster R-CNN~\cite{faster2015} or the single shot multibox detector (SSD)~\cite{ssd2016}. 
Baweja et al.~\cite{baweja2018stalknet} used Faster R-CNN to count plant stalks and calculate the stalk width from on-site images. 
Faster R-CNN has also been utilized to localize the plucking points of tea leaves~\cite{chen2018} to support automated tea harvesting. 
Fuentes et al.~\cite{fuentes2017robust} trained three different CNN architectures for the real-time localization and diagnosis of tomato diseases. 
They attained a maximum mean average precision of 86.0\% in the diagnosis of a total of nine types of tomato diseases and pests. 
They also proposed a CNN filter bank to reduce the false positive rate for the detection of tomato diseases caused by the data class imbalance problem~\cite{fuentes2018high}. 
They reported a mean average precision of 96.0\% for nine diseases and one physiological disorder diagnosis for tomatoes. 
In another study on the diagnosis of in-field wheat diseases, Lu et al.~\cite{lu2017field} designed an end-to-end diagnosis system for wheat by applying fully convolutional neural networks, and reported a mean recognition accuracy of 98.0\% for the classification of seven classes (six diseases and healthy class).

Although these proposals demonstrated the excellent adaptability of the deep learning approach in terms of performing plant disease detection and diagnosis from real scenario images, most of these systems were designed to diagnose a limited number of targets (e.g. up to a dozen), meaning that there are still limitations when applying these methods to large-scale farm environments. 
A system that can accurately detect and diagnose diseases from wide-angle images is extremely important in order to support agricultural practice.

However, based on our experience, it is not easy to develop a practical plant disease diagnosis system for wide-angle images. 
There are two major problems that need special attention. 
The first problem is the overfitting that arises due to the latent similarities between the training and test images, even though they were exclusive to each other. 
This typically happens when images photographed in the same field are divided into training and test sets. 
This scenario is commonly seen (i.e. performance evaluation is done using cross-validation), and the diagnostic performance on real unseen data is usually significantly reduced. 
To make matters worse, this problem is predicted to be more serious when wide-angle images are used, because the same or similar objects may appear in different images.

The second problem is the labor cost and the required accuracy of the gold standard assignment. 
When an end-to-end diagnosis system is built, numerous training images with a huge number of bounding boxes are required, along with the appropriate disease labels. 
Moreover, there are innumerable objects in the images such as overlapping leaves, and their resolution is often insufficient, making it very difficult to label each object with an appropriate ground truth.

Currently, there are very few attempts to diagnose plant diseases from in-field wide-angle images. 
Cap et al.~\cite{quan2018} were the first to propose a cucumber diagnosis system using wide-angle on-site images. 
They developed a two-stage system consisting of two CNNs that performed leaf detection and leaf diagnosis independently. 
Their system was evaluated under practical settings and achieved a reasonable performance. 
However, they have not tested their system on images taken from different environments (e.g. different farms) nor compared with other state-of-the-art end-to-end techniques (e.g. Faster R-CNN or SSD).

Nevertheless, we believe that a two-stage diagnosis system such as the one in~\cite{quan2018} has several advantages that can overcome the issues on developing practical plant disease diagnosis system for wide-angle images. 
Firstly, two-stage systems have the detection stage and the diagnosis stage separately; thus, labeling of the bounding boxes of the objects (i.e. leaves, fruits) to be detected is easy, since it does not require disease-specific knowledge and can be done by non-experts. 
Secondly, labeling a single object or collecting labeled single-object images is much simpler than for wide-angle images, as mentioned above. 
The diagnosis stage therefore could be trained with a wider variety of data, boosting the robustness of the two-stage system when new types of data are encountered. 
In this paper, we examine the difficulties involved in realizing the automatic diagnosis of plant diseases using practical wide-angle images and propose a suitable configuration for this problem.

\input{tables/table_1.tex}

\begin{figure*}[hbt!]
\centering
\includegraphics[width=0.85\linewidth]{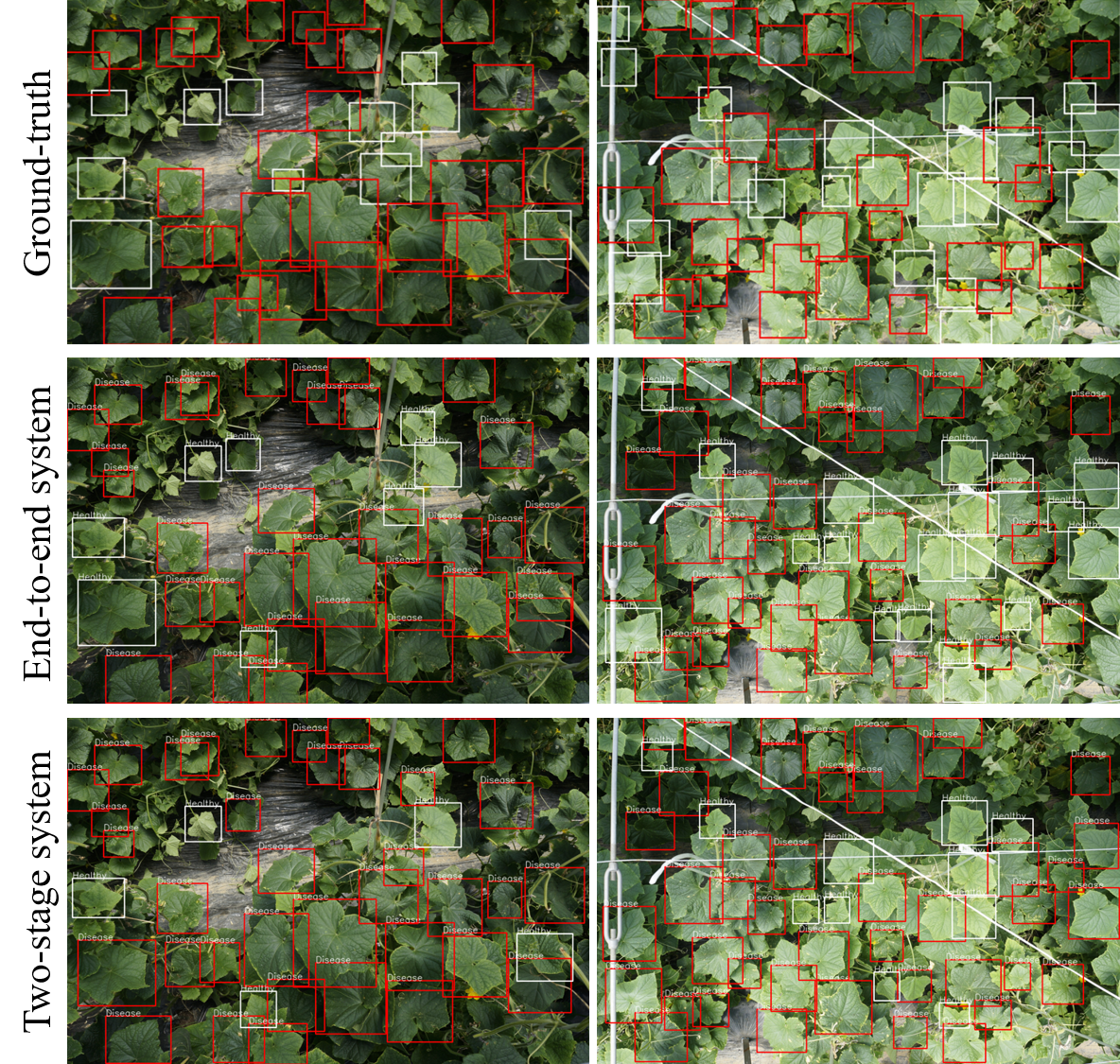}
\caption{
    Final diagnostic result of two diagnosis strategies on the $\text{wide-angle}_\text{test}$ dataset. 
    The first row represents the ground-truth images, and the second and third rows indicate the results of the end-to-end SSD512 system and the two-stage system with $\text{DiagNet}_\text{all}$, respectively. 
    Note that the red and white boxes show diseased and healthy cases, respectively.
}
\label{fig:examples_of_test_dataset}
\end{figure*}

%% file: tables/table_1.tex
\begin{table*}[hbt!]
\centering

\caption{Performance comparison between end-to-end and two-stage systems on the $\text{wide-angle}_\text{test}$ dataset}
\label{tab:table_1}
\resizebox{\textwidth}{!}{%
\begin{threeparttable}
\begin{tabular}{@{}cccccccccccc@{}}
\toprule
\multicolumn{2}{c}{}                                                         & \multicolumn{3}{c}{Leaf detector performance}                                                                                                                              & \multicolumn{7}{c}{Leaf disease diagnostic performance}                                                                                                                                                                                                                                                                            \\ \cmidrule(lr){3-5}  \cmidrule(lr){6-12}
\multicolumn{2}{c}{}                                                         & \multicolumn{1}{r}{}                                    & \multicolumn{1}{l}{}                                     & \multicolumn{1}{l}{}                                  & \multicolumn{3}{c}{Healthy}                                                                                          & \multicolumn{3}{c}{Disease}                                                                                         &                                                                                       \\ \cmidrule(lr){6-8} \cmidrule(lr){9-11}
\multicolumn{2}{c}{\multirow{-3}{*}{}}                                       & \multicolumn{1}{r}{\multirow{-2}{*}{F1-score {[}\%{]}}} & \multicolumn{1}{l}{\multirow{-2}{*}{Precision {[}\%{]}}} & \multicolumn{1}{l}{\multirow{-2}{*}{Recall {[}\%{]}}} & \multicolumn{1}{l}{F1-score {[}\%{]}} & \multicolumn{1}{l}{Precision {[}\%{]}} & \multicolumn{1}{l}{Recall {[}\%{]}} & F1-score {[}\%{]}                    & \multicolumn{1}{l}{Precision {[}\%{]}} & \multicolumn{1}{l}{Recall {[}\%{]}} & \multirow{-2}{*}{\begin{tabular}[c]{@{}c@{}}Average\\ F1-score {[}\%{]}\end{tabular}} \\ \cmidrule(lr){1-2} \cmidrule(lr){3-5} \cmidrule(lr){6-8} \cmidrule(lr){9-11} \cmidrule(lr){12-12}
                               & End-to-end                                  &                                                         &                                                          &                                                       & {\color[HTML]{FE0000} \textbf{87.8}}  & 88.1                                   & 87.5                                & {\color[HTML]{FE0000} \textbf{84.1}} & 81.4                                   & 86.9                                & {\color[HTML]{FE0000} \textbf{86.0}}                                                  \\
                               & Two-stage ($\text{DiagNet}_\text{all}$)     &                                                         &                                                          &                                                       & 82.6                                  & 88.0                                   & 77.9                                & 73.2                                 & 62.8                                   & 87.6                                & 77.9                                                                                  \\
\multirow{-3}{*}{SSD512}       & Two-stage ($\text{DiagNet}_\text{cropped}$) & \multirow{-3}{*}{\textbf{91.5}}                         & \multirow{-3}{*}{89.8}                                   & \multirow{-3}{*}{93.3}                                & {\color[HTML]{3531FF} \textbf{84.6}}   & 86.0                                   & 83.3                                & {\color[HTML]{3531FF} \textbf{79.1}} & 75.0                                   & 83.7                                & {\color[HTML]{3531FF} \textbf{81.9}}                                                  \\ \cmidrule(lr){1-1} \cmidrule(lr){2-2} \cmidrule(lr){3-5} \cmidrule(lr){6-8} \cmidrule(lr){9-11} \cmidrule(lr){12-12}
                               & End-to-end                                  &                                                         &                                                          &                                                       & 85.2                                  & 82.8                                   & 87.8                                & 81.5                                 & 78.7                                   & 84.6                                & 83.4                                                                                  \\
                               & Two-stage ($\text{DiagNet}_\text{all}$)     &                                                         &                                                          &                                                       & 80.8                                  & 81.5                                   & 80.1                                & 75.1                                 & 68.3                                   & 83.5                                & 78.0                                                                                  \\
\multirow{-3}{*}{Faster R-CNN} & Two-stage ($\text{DiagNet}_\text{cropped}$) & \multirow{-3}{*}{90.4}                                  & \multirow{-3}{*}{86.7}                                   & \multirow{-3}{*}{94.4}                                & 82.8                                  & 81.6                                   & 84.1                                & 77.7                                 & 73.2                                   & 82.9                                & 80.3                                                                                  \\ \bottomrule
\end{tabular}%
\begin{tablenotes}
\item[]The {\color[HTML]{FE0000} red} and {\color[HTML]{3531FF} blue} colors indicate the best performance of the end-to-end and two-stage systems on the $\text{wide-angle}_\text{test}$ dataset, respectively.
\end{tablenotes}
\end{threeparttable}
}
\end{table*}

%% file: 02_method.tex
We examine and compare two types of diagnosis strategy for practical wide-angle cucumber images in terms of disease diagnostic performance under different evaluation environments. 
\figref{fig:the_overview_system} shows an overview of the two different types of diagnosis strategy. 
The first approach is an end-to-end strategy, which simultaneously performs leaf detection and diagnosis based on a sophisticated object detection framework such as the SSD or Faster R-CNN. 
The second approach is a two-stage strategy that performs these functions separately. 
In this study, we carry out diagnosis using these strategies in order to estimate whether each leaf in a wide-angle image is healthy or diseased. 
The reason for using only two diagnostic classes is that it is difficult to assign gold standard labeling to wide-angle images, as described earlier. 
For both systems, we compare the final diagnostic performance between a test dataset from the same farm and datasets from different farms. 
This comparison is to examine the effect of the latent similarities between the training and test datasets on the final diagnostic performance. 
We then discuss which approach is more suitable for real cultivation conditions.

\subsection{Datasets}
In this work, we use the following two datasets to explore suitable configurations for automatic wide-angle diagnosis of plant disease.

\subsubsection{Wide-angle dataset}
A total of 963 wide-angle images of cucumbers were acquired from several farms, using various digital cameras. 
Each wide-angle image contained numerous cucumber leaves that overlapped each other and was taken under different light conditions (see \figref{fig:the_overview_system}--3 for sample images). 
The images contained a total of 24,565 leaves, of which 16,924 were healthy and 7,641 were diseased. 
All of the wide-angle images were annotated by experts, and bounding boxes were created for each leaf. 
We randomly divided the images, using 90\% (867 wide-angle images containing 15,369 healthy and 6,827 diseased leaves) for training, and the rest for testing (96 wide-angle images containing 1,555 healthy and 814 diseased leaves). 
We refer to these sets of images as the $\text{wide-angle}_\text{train}$ and the $\text{wide-angle}_\text{test}$ datasets, respectively.

In order to evaluate the end-to-end systems and the two-stage systems equally, we prepared 51 wide-angle cucumber images taken from completely different farms. 
A total of 1,829 single leaves (of which 820 are healthy and 1,009 diseased) were also annotated by experts. 
We used this dataset only for the final diagnostic test and refer to it here as the $\text{wide-angle}_\text{unseen}$ dataset.

Our wide-angle images mainly had two aspect ratios, 2:3 and 3:4, and the typical resolution of these images was between 12 and 20 megapixels. 
They were resized to either 512$\times$512, 600$\times$900 or 600$\times$800 pixels, depending on the architecture of the end-to-end models (as described in more detail in the experimental section).

\subsubsection{Single-leaf dataset}
The single-leaf dataset was used for training the diagnosis stage of the two-stage systems. 
From the 867 images in the $\text{wide-angle}_\text{train}$ dataset, we cropped all the gold standard bounding boxes (a total of 22,196 images, 15,369 healthy and 6,827 diseased), each containing one cucumber leaf, to form the dataset. 
We refer this dataset as the $\text{single-leaf}_\text{cropped}$ dataset. 
In addition, we combined these images with another set of single-leaf images collected from Saitama Agricultural Technology Research Center, Japan. 
Note that these images were not included in the abovementioned wide-angle dataset. 
This formed the $\text{single-leaf}_\text{all}$ dataset, which contains 50,000 images of single cucumber leaves (25,000 healthy and 25,000 diseased).

The reason for building this larger $\text{single-leaf}_\text{all}$ dataset is to verify the advantages of the two-stage systems, as hypothesized earlier in the introduction. 
The end-to-end systems only accept annotated leaf regions in wide-angle images in the training set, while the two-stage systems could include additional single-leaf images in the training of the diagnosis stage. 
Note again that the acquisition of labeled single-leaf images is much easier than from wide-angle images. 
We believe that adding a variety of single-leaf images to the dataset can improve the robustness of the diagnostic model. 
The resolution of the single-leaf dataset was normalized to 224$\times$224 pixels.

\subsection{Wide-angle plant diagnosis systems}
\subsubsection{End-to-end systems}
We first built our end-to-end systems using the SSD512 and Faster R-CNN models. 
The input image size was resized to 512$\times$512 pixels for the SSD model, while for the Faster R-CNN, we resized the input images to sizes 600$\times$900 or 600$\times$800 pixels, corresponding to images with aspect ratios of 2:3 or 3:4. 
The backbone of these models was basically the VGG-16~\cite{vgg2014} model pre-trained with the ImageNet dataset~\cite{deng2009imagenet}, and they were fine-tuned with the $\text{wide-angle}_\text{train}$ dataset. 
The diagnostic performance of the end-to-end systems was evaluated and compared on the $\text{wide-angle}_\text{test}$ and $\text{wide-angle}_\text{unseen}$ datasets.

\subsubsection{Two-stage systems}
A two-stage system is a combination of a leaf detection stage and a leaf diagnosis stage. 
In the leaf detection stage, we used the above end-to-end systems (i.e. SSD512 or Faster R-CNN) as the leaf detectors to enable an unbiased comparison. 
In the subsequent leaf diagnosis stage, the detected leaves were diagnosed using an additional CNN model called DiagNet. 
We used the CNN model proposed in~\cite{quan2018} for the performance comparison. 
This network was also a fine-tuned version of the pre-trained VGG-16 network with two outputs, i.e. healthy or diseased. 
Our DiagNet model accepts a color image with a size of 224$\times$224 pixels. 
In this work, we froze the first ten convolutional layers and fine-tuned the last six layers (three convolutional and three fully-connected layers).

For experimental purposes, we trained two versions of the DiagNet model for performance comparison. 
The first model, named $\text{DiagNet}_\text{cropped}$, was trained only on the $\text{single-leaf}_\text{cropped}$ dataset (22,196 images), while the other, called $\text{DiagNet}_\text{all}$, was trained on the $\text{single-leaf}_\text{all}$ dataset (50,000 images). 
The diagnostic performance of the two-stage systems was also evaluated and compared using the $\text{wide-angle}_\text{test}$ and $\text{wide-angle}_\text{unseen}$ datasets.

\subsubsection{Training wide-angle plant diagnosis systems}
To train the end-to-end systems, the Faster R-CNN and SSD512 models were fine-tuned using the $\text{wide-angle}_\text{train}$ dataset. 
We followed the training strategy used in the original Faster R-CNN and SSD papers, fine-tuning the models using stochastic gradient descent (SGD) with momentum~\cite{qian1999momentum} with an initial learning rate of $10^{-3}$ a momentum of $0.9$, and a weight decay of $0.0005$. 
The mini-batch size was set to one to train the Faster R-CNN and 16 to train the SSD512. 
The training was terminated after 50,000 iterations.

For the two-stage systems, the $\text{DiagNet}_\text{cropped}$ and $\text{DiagNet}_\text{all}$ were trained on the $\text{single-leaf}_\text{cropped}$ and $\text{single-leaf}_\text{all}$ datasets, respectively. 
During the training, we applied augmentation on the fly, using horizontal and vertical flipping, and random 90 degrees rotations. 
We used the SGD momentum optimizer with the same hyper-parameters when training both the end-to-end systems and our two-stage models. 
The minibatch size was set to 256, and we terminated the training process after 30 epochs.

%% file: 03_experiment.tex
We compare the diagnostic performance of the two different diagnosis strategies for the wide-angle pictures taken on the same farm and those from different farms. 
Again, it should be noted here that the purpose of this experiment is to find a suitable configuration for practical systems based on this comparison. 
More specifically, we clarify the effect of the latent similarities in the dataset, and propose a suitable solution to this problem. 
In this experiment, diagnosis bounding boxes with an intersection over union (IoU) $\geq0.5$ which corresponding to the ground-truth label are regarded as correct detection results. 
We use the evaluation criteria of precision, recall and F1-score for both healthy and diseased cases, and calculate the average diagnostic F1-score by averaging the F1-scores of the healthy and diseased leaves as an indicator of the overall diagnostic performance.

\subsubsection{Experiment 1: Diagnosing the $\text{wide-angle}_\text{test}$ dataset}
\tabref{tab:table_1} shows a comparison of the performance in terms of leaf detection and leaf diagnosis on the $\text{wide-angle}_\text{test}$ dataset (96 images, containing 1,555 healthy and 814 diseased leaves). 
These results show that the best leaf detection performance is achieved by SSD512 with an F1-score of 91.5\%, which is slightly better than the Faster R-CNN with 90.4\%. 
The diagnostic results show that the end-to-end systems give better performance on diagnosing diseased leaves compared to the two-stage systems. 
The best average diagnostic F1-score is 86.0\% for the SSD512, while the best result for the two-stage systems is 81.9\% for the $\text{DiagNet}_\text{cropped}$ using SSD512 as the leaf detector. 
The overall ranking indicates that of the end-to-end systems, the SSD512 performed slightly better than the Faster R-CNN. 
For the two-stage systems, the $\text{DiagNet}_\text{cropped}$ achieved higher results than the $\text{DiagNet}_\text{all}$ using both SSD512 and Faster R-CNN as the leaf detectors. 
We should note here that the $\text{DiagNet}_\text{all}$ was trained with a much larger leaf image dataset (roughly 2.3 times larger than the $\text{DiagNet}_\text{cropped}$), but the performance was consistently lower.

\figref{fig:examples_of_test_dataset} shows some examples from this experiment. 
The first row represents the ground truth images, while the second and third rows represent the results from the end-to-end and two-stage systems, respectively. 
The red and white boxes indicate diseased and healthy leaves, respectively. 
Based on the results, it can be seen that although there is a slight difference in the performance for the two types of system, both the end-to-end and two-stage systems can correctly diagnose almost all leaf locations, giving reasonable diagnostic performance.

\subsubsection{Experiment 2: Diagnosing the $\text{wide-angle}_\text{unseen}$ dataset}
\tabref{tab:table_2} shows a comparison of the performance of leaf detection and diagnosis for the $\text{wide-angle}_\text{unseen}$ dataset, which contains 51 images (820 healthy and 1,009 diseased leaves). 
Again, these images were taken in a completely different environment from the above $\text{wide-angle}_\text{test}$ dataset.

The leaf detection performance for these unseen images was significantly reduced with respect to the recall, but both SSD and Faster RCNN maintained a very high value of precision (94.4 -- 96.1\%). 
From a practical point of view, this can be considered reasonable, since we still can detect most leaves precisely. 
The best leaf detection performance in this case is achieved by the Faster R-CNN model, with an F1-score of 54.2\% as compared to the SSD512 with 51.8\%. 
The final diagnostic performance was totally dissimilar from the previous experiment, as showed in \tabref{tab:table_1}. 
Although all systems showed a considerably reduced diagnostic performance, the two-stage systems outperformed the end-to-end systems. 
The best average diagnostic F1-score for the two-stage systems is 37.3\% for the $\text{DiagNet}_\text{all}$, while the best end-to-end system is the Faster R-CNN diagnostic system with only 20.7\%. 
It is notable that both the SSD512 and Faster R-CNN end-to-end systems were almost unable to detect the locations of diseased leaves, with a very low F1-score of 4.4 -- 6.2\%.

In contrast, the two-stage system ($\text{DiagNet}_\text{all}$) achieved much higher recall and precision for the diseased cases, attaining F1-score of 33.4 -- 38.9\%. 
The diagnostic performance of $\text{DiagNet}_\text{all}$ was also well balanced between the healthy and diseased cases. 
Along with that, the $\text{DiagNet}_\text{cropped}$ still attained a desirable result in terms of diagnosing disease, even with a smaller set of training data. 
The overall performance ranking is opposite to that in the previous experiment, with the best result achieved by $\text{DiagNet}_\text{all}$, the second best by $\text{DiagNet}_\text{cropped}$, and the lowest by the end-to-end systems.

\begin{figure*}[hbt!]
\centering
\includegraphics[width=0.85\linewidth]{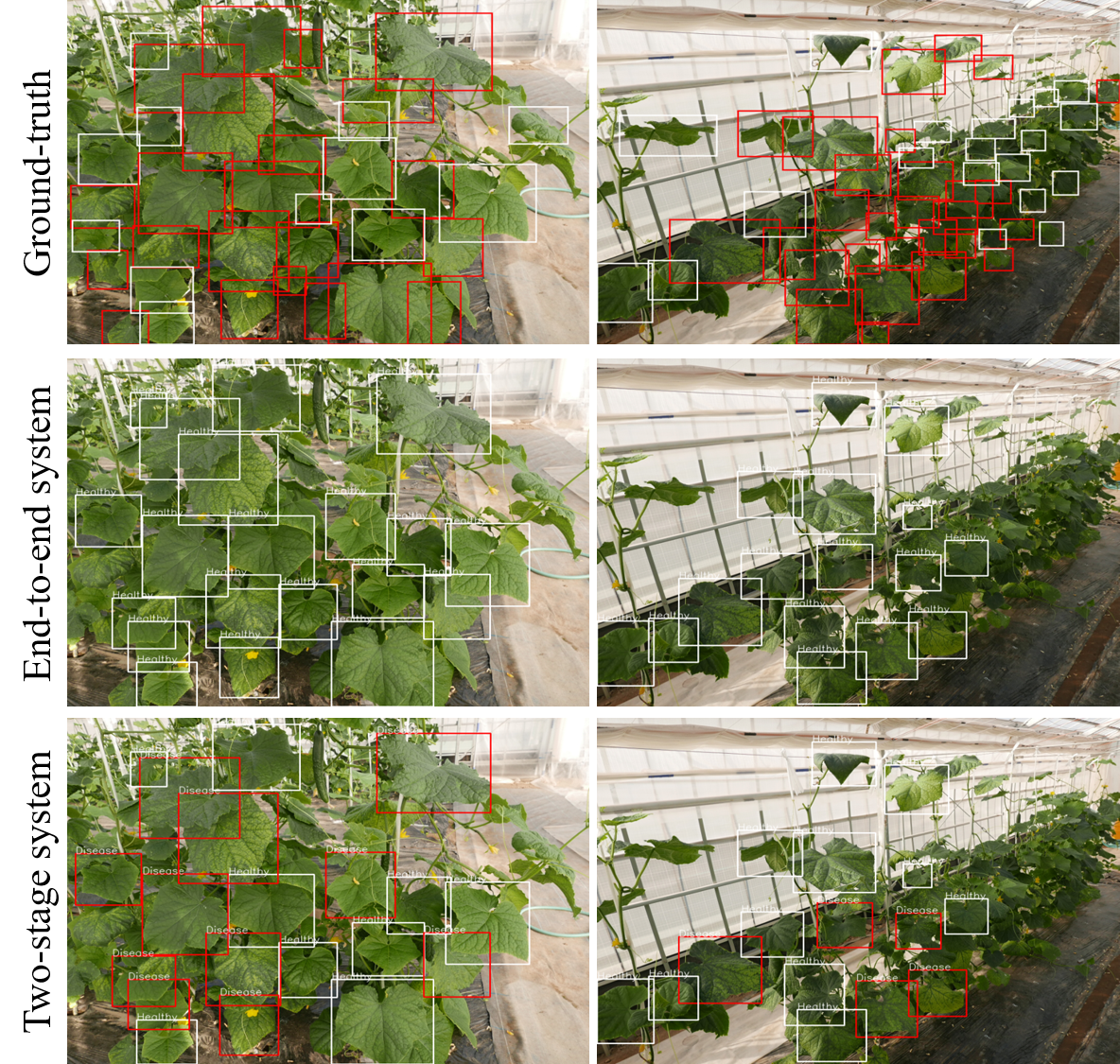}
\caption{
    Final diagnostic results of two strategies on the $\text{wide-angle}_\text{unseen}$ dataset. 
    The first row shows the ground-truth images, while the second and third rows indicate the results of the end-to-end Faster R-CNN system and the two-stage system with $\text{DiagNet}_\text{all}$, respectively. 
    The end-to-end system completely failed to detect the diseased leaves, while the two-stage system correctly diagnosed the important leaf locations in the unseen dataset.
}
\label{fig:examples_of_unseen_datase}
\end{figure*}

\figref{fig:examples_of_unseen_datase} shows typical examples of the final diagnostic results for the $\text{wide-angle}_\text{unseen}$ dataset. 
As mentioned above, the end-to-end systems typically failed to diagnose the positions of diseased leaves, while the two-stage systems could correctly identify the important locations of diseased leaves for an unseen dataset, outperforming the end-to-end systems. 

\input{tables/table_2.tex}

%% file: tables/table_2.tex
\begin{table*}[hbt!]
\centering
\caption{Performance comparison between end-to-end and two-stage systems on the $\text{wide-angle}_\text{unseen}$ dataset}
\label{tab:table_2}
\resizebox{\textwidth}{!}{%
\begin{threeparttable}
\begin{tabular}{@{}cccccccccccc@{}}
\toprule
\multicolumn{2}{c}{}                                                         & \multicolumn{3}{c}{Leaf detector performance}                                                                                                                              & \multicolumn{7}{c}{Leaf disease diagnostic performance}                                                                                                                                                                                                                                                                            \\ \cmidrule(lr){3-5}  \cmidrule(lr){6-12}
\multicolumn{2}{c}{}                                                         & \multicolumn{1}{r}{}                                    & \multicolumn{1}{l}{}                                     & \multicolumn{1}{l}{}                                  & \multicolumn{3}{c}{Healthy}                                                                                          & \multicolumn{3}{c}{Disease}                                                                                         &                                                                                       \\ \cmidrule(lr){6-8} \cmidrule(lr){9-11}
\multicolumn{2}{c}{\multirow{-3}{*}{}}                                       & \multicolumn{1}{r}{\multirow{-2}{*}{F1-score {[}\%{]}}} & \multicolumn{1}{l}{\multirow{-2}{*}{Precision {[}\%{]}}} & \multicolumn{1}{l}{\multirow{-2}{*}{Recall {[}\%{]}}} & \multicolumn{1}{l}{F1-score {[}\%{]}} & \multicolumn{1}{l}{Precision {[}\%{]}} & \multicolumn{1}{l}{Recall {[}\%{]}} & F1-score {[}\%{]}                    & \multicolumn{1}{l}{Precision {[}\%{]}} & \multicolumn{1}{l}{Recall {[}\%{]}} & \multirow{-2}{*}{\begin{tabular}[c]{@{}c@{}}Average\\ F1-score {[}\%{]}\end{tabular}} \\ \cmidrule(lr){1-2} \cmidrule(lr){3-5} \cmidrule(lr){6-8} \cmidrule(lr){9-11} \cmidrule(lr){12-12}
                               & End-to-end                                  &                                                         &                                                          &                                                       & 34.5                                  & 39.4                                   & 30.7                                & 4.4                                  & 66.7                                   & 2.3                                 & 19.5                                                                                  \\
                               & Two-stage ($\text{DiagNet}_\text{all}$)     &                                                         &                                                          &                                                       & 36.2                                  & 53.9                                   & 27.2                                & 33.4                                 & 81.2                                   & 21.0                                & 34.8                                                                                  \\
\multirow{-3}{*}{SSD512}       & Two-stage ($\text{DiagNet}_\text{cropped}$) & \multirow{-3}{*}{51.8}                                  & \multirow{-3}{*}{96.1}                                   & \multirow{-3}{*}{35.5}                                & 35.4                                  & 44.1                                   & 29.6                                & 17.4                                 & 80.6                                   & 9.9                                 & 26.4                                                                                  \\ \cmidrule(lr){1-1} \cmidrule(lr){2-2} \cmidrule(lr){3-5} \cmidrule(lr){6-8} \cmidrule(lr){9-11} \cmidrule(lr){12-12}
                               & End-to-end                                  &                                                         &                                                          &                                                       & {\color[HTML]{FE0000} \textbf{35.1}}  & 38.2                                   & 32.4                                & {\color[HTML]{FE0000} \textbf{6.2}}  & 84.6                                   & 3.2                                 & {\color[HTML]{FE0000} \textbf{20.7}}                                                  \\
                               & Two-stage ($\text{DiagNet}_\text{all}$)     &                                                         &                                                          &                                                       & {\color[HTML]{3531FF} \textbf{35.6}}  & 53.2                                   & 26.7                                & {\color[HTML]{3531FF} \textbf{38.9}} & 79.9                                   & 25.7                                & {\color[HTML]{3531FF} \textbf{37.3}}                                                  \\
\multirow{-3}{*}{Faster R-CNN} & Two-stage ($\text{DiagNet}_\text{cropped}$) & \multirow{-3}{*}{\textbf{54.2}}                         & \multirow{-3}{*}{94.4}                                   & \multirow{-3}{*}{38.0}                                & 34.7                                  & 40.4                                   & 30.4                                & 15.9                                 & 75.0                                   & 8.9                                 & 25.3                                                                                  \\ \bottomrule
\end{tabular}%
\begin{tablenotes}
\item[] {The \color[HTML]{FE0000} red} and {\color[HTML]{3531FF} blue} colors indicate the best performance of end-to-end and two-stage systems on the $\text{wide-angle}_\text{unseen}$ dataset, respectively.
\end{tablenotes}
\end{threeparttable}
}
\end{table*}

%% file: 04_discussion.tex
We investigated changes in diagnostic performance by experimenting with different practical scenarios, and have shown that the final diagnostic performance varies greatly depending on whether the test data form part of the whole dataset (\tabref{tab:table_1}; Experiment 1) or a completely different dataset (\tabref{tab:table_2}; Experiment 2). 
The results of both experiments indicated that the end-to-end systems were overfitted to the $\text{wide-angle}_\text{train}$ dataset. 
The end-to-end Faster R-CNN and SSD512 models showed very high performance for disease diagnosis on the $\text{wide-angle}_\text{test}$ dataset (F1-score 81.5 -- 84.1\%) but extremely poor performance on the $\text{wide-angle}_\text{unseen}$ dataset (F1-score 4.4 -- 6.2\%). 
The primary reason for this huge gap is the large latent similarities between the training and test data (i.e. there is a high possibility that the same or a similar object appears in the wide-angle images in the same field.). 
In addition, collecting a sufficiently large and reliable wide-angle training dataset is difficult even for experts, because the leaf objects that need to be labeled are often small, with unclear appearance. 
This limits the scalability of the system, leading to non-generalization to the unseen dataset. 
In this case, the end-to-end systems are not the best choice for practical automated disease diagnosis.

In contrast, although the two-stage systems attained a slightly lower F1-score than the end-to-end systems in Experiment 1, they showed superior performance in diagnosing disease cases from the $\text{wide-angle}_\text{unseen}$ dataset, which represented a more practical scenario in Experiment 2 (with an F1-score of 33.4 -- 38.9\% compared to 4.4 -- 6.2\% for the end-to-end systems). 
We also showed that even when we used only the cropped single-leaf images from the same training dataset (i.e. $\text{single-leaf}_\text{cropped}$ dataset is cropped from the $\text{wide-angle}_\text{train}$ dataset) as in the end-to-end systems, $\text{DiagNet}_\text{cropped}$ still achieved better results for diagnosing an unseen disease, with an F1-score ranging from 15.9\% to 17.4\%, thus confirming the effectiveness of the two-stage strategy in real situations.

We achieved these results thanks to the advantages of the training method in the two-stage strategy. 
First, the leaf diagnosis stage in a two-stage system accepts single-leaf images as input, and can be trained with a wide variety of data. 
Following this, the collection of single-leaf images for the disease classifier is much more straightforward than for end-to-end systems (i.e. wide-angle images). 
These two properties therefore contribute to improving the generalization of the two-stage systems and increasing their scalability.

In general, although the detection of the location of a diseased leaf is very important, it is unnecessary to accurately detect and diagnose all the leaves. 
Once the areas of diseased leaves are detected, further inspection can be applied to the nearby locations, since plant diseases often spread outwards from a given area. 
In this context, a two-stage system is a suitable choice for the diagnosis of plant diseases from wide-angle images in practical situations.

%% file: 05_conclusion.tex
In this paper, we have explored the difficulties of establishing practical plant disease diagnosis systems for wide-angle images, and have compared two diagnosis strategies to solve these issues. 
Our experiments demonstrate that even sophisticated end-to-end systems such as Faster R-CNN and SSD still fall into overfitting and cannot achieve the desired performance for an unknown dataset. 
On the other hand, although they require further improvement, our two-stage systems attained promising disease diagnosis performance for the unseen target dataset.

These results show that it is preferable to use two-stage systems due to the greater ease of collecting training data and assigning ground-truth labels, and due to the performance improvement they give. 
In future work, we intend to improve our models with data on more types of disease to give better performance.